\documentclass{article}

\PassOptionsToPackage{numbers,sort&compress}{natbib}

 \usepackage[dblblindworkshop, final]{neurips_2026}
\workshoptitle{AXIOM Foundations of Efficient Deep Learning}

\usepackage{amsmath,amsfonts,bm}

\newcommand{\chinOne}{L = E + AC^\alpha}

\newcommand{\chinThree}{L(N, D) = E + \frac{A}{N^\alpha} + \frac{B}{D^\beta}}

\def\1{\bm{1}}

\DeclareMathAlphabet{\mathsfit}{\encodingdefault}{\sfdefault}{m}{sl}
\SetMathAlphabet{\mathsfit}{bold}{\encodingdefault}{\sfdefault}{bx}{n}

\DeclareMathOperator*{\argmin}{arg\,min}

\usepackage{accents}
\newlength{\dhatheight}

\usepackage[utf8]{inputenc} 
\usepackage[T1]{fontenc}    
\usepackage{hyperref}       
\hypersetup{hypertexnames=false} 
\usepackage{url}            
\usepackage{booktabs}       
\usepackage{amsfonts}       
\usepackage{nicefrac}       
\usepackage{microtype}      
\usepackage{xcolor}         
\usepackage[colorinlistoftodos]{todonotes}
\usepackage{algorithm}
\usepackage{algpseudocode}
\usepackage{amsmath,amssymb}
\usepackage{graphicx}
\usepackage{longtable}

\usepackage{paralist}       
\usepackage[capitalize,noabbrev]{cleveref}  
\usepackage[most]{tcolorbox}
\usepackage{multirow}

\algrenewcommand\algorithmicrequire{\textbf{Input:}}
\algrenewcommand\algorithmicensure{\textbf{Output:}}

\definecolor{dianacolor}{HTML}{1F4E79}

\title{Amortizing Scaling Law Construction Costs}

\makeatletter
\newcommand{\unmarked@footnotetext}[1]{%
  \begingroup
    \renewcommand{\thefootnote}{}%
    \long\def\@makefntext##1{\parindent 1em\noindent ##1}%
    \footnotetext{#1}%
  \endgroup
}
\newcommand{\unmarkedthanks}[1]{%
  \protected@xdef\@thanks{\@thanks\protect\unmarked@footnotetext{#1}}%
}
\makeatother

\author{%
  Abhash Kumar Jha\textsuperscript{1,3,*}
  \unmarkedthanks{Correspondence to: \texttt{\{abhash.kumar.jha,aaron.klein\}@tue.ellis.eu}} \And
  Diana Alexandra Onu\textcommabelow{t}u\textsuperscript{2,3,*} \And
  Neeratyoy Mallik\textsuperscript{4,5,1,6,*}
  \AND
  Swagatam Haldar\textsuperscript{1,3,*} \And
  Sam Laing\textsuperscript{1,3} \And
  Niccolò Ajroldi\textsuperscript{1,3} \And
  Shiwei Liu\textsuperscript{1,6,7} 
  \AND
  Joaquin Vanschoren\textsuperscript{2,3} \qquad
  Aaron Klein\textsuperscript{1,3}
  \\
  \\
  \textsuperscript{1}ELLIS Institute Tübingen \quad
  \textsuperscript{2}Eindhoven University of Technology \\
  \textsuperscript{3}OpenEuroLLM \quad
  \textsuperscript{4}University of Freiburg \quad
  \textsuperscript{5}Zuse School ELIZA \quad \\
  \textsuperscript{6}Max Planck Institute for Intelligent Systems \quad
  \textsuperscript{7}Tübingen AI Center \\
  \textsuperscript{*}Equal Contribution. 
}

\begin{document}

\maketitle

\begin{abstract}

Scaling laws guide the design choices for training large foundation models, but deriving them involves training an exhaustive grid over hyperparameters, token budgets, and parameter counts, which is computationally expensive. 
Fitting a scaling law, however, only requires the best-loss frontier across compute scales, discarding most of the trained configurations.
We propose a framework for efficient scaling law construction that formulates data collection as a Bayesian optimization problem, and introduce metrics for comparing scaling law fitting methods under constrained compute budgets.
We find that progressively expanding the compute budget during acquisition, mirroring the compute-ordered evaluation of configurations in practice, substantially improves recovery efficiency. 
Augmenting the observed configurations with surrogate-fantasized evaluations then recovers the broader experimental grid, allowing accurate scaling law fitting without training every configuration. 
Together, these can closely match scaling law fits over a full dense grid at computational savings of up to $10\text{--}100\times$. Code is open-sourced on \href{https://github.com/OpenEuroLLM/amortizing-scaling-law-fitting-costs}{GitHub}. 

\end{abstract}

\section{Introduction}\label{sec:intro}










Pretraining large language models requires decisions for a wide range of design choices. 
However, evaluating each alternative of model architecture, data mixture, optimization settings at the target scale is prohibitively expensive.
Scaling laws~\citep{kaplan-arxiv20a,hoffmann-neurips22a,alabdulmohsin2023getting} allow us to extrapolate model performance and optimal training configurations as we scale, by fitting a power law between quantities of interest, such as loss or hyperparameters, and the compute budget~\citep{clark-icml22a,li-neurips24a,bjorck-iclr25a,li-arxiv25a,bergsma2026power,yin-cvpr25a,zhang-iclr25a}. 
To derive them, we train a predefined dense grid of configurations across various scales.
This requires careful experimental design of the grid over hyperparameters, token budgets, and parameter counts.

Classical hyperparameter optimization (HPO)~\citep{frazier-arxiv18a,feurer-automlbook19a,franceschi-ftml25} automatically selects the optimal hyperparameters that minimize the validation loss, typically at a single model or data scale. 
Multi-fidelity HPO~\citep{jamieson-aistats16a,poloczek-nips17a,falkner-icml18a,astudillo-arxiv22a} and learning-curve extrapolation~\citep{domhan-ijcai15a,klein-iclr17a,adriaensen2023efficient,pmlr-v235-rakotoarison24a} optimize over data scales, but are usually associated with early stopping of individual configurations. 
Parameter count, meanwhile, can also be seen as a fidelity affecting compute cost~\citep{carstensen-automl25a}, and can interact with data scale to drive hyperparameter drift~\citep{yang-neurips21a}. 
Tuning for a best-loss envelope for scaling law construction is therefore markedly different from the usual flavours of HPO, which return a single best-loss configuration.

\begin{figure}
    \centering
    \includegraphics[width=0.95\linewidth]{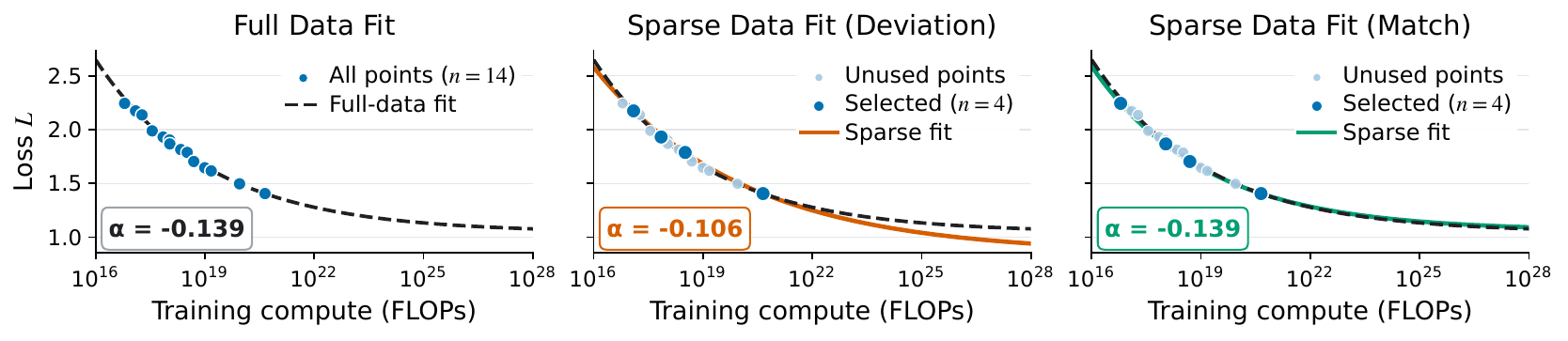}
    \caption{
        Scaling law $\chinOne{}$ fit to different loss subsets.  \textit{Left}: the full Pareto envelope based all 14 different observations and its fitted law;  
        \textit{Middle}: a poorly selected four-point subset produces a deviated fit;   
        \textit{Right}: a different 4-point subset closely recovers the full-data fit. 
        Both sparse fits use the same number of observations, but their accuracy depends strongly on which points are selected, motivating the problem of finding a small subset that reliably recovers the full scaling law.
   }\label{fig:fig1-pitch}
\end{figure}

A parametric scaling law fit is a standard regression task where the data collection procedure directly affects the quality of the fit (see~\Cref{fig:fig1-pitch}). For each point in the parametric form, we use only a subset consisting of the minimum loss over the independent variables. This provides a direct source of efficiency gains in data collection for scaling law fitting. Since only the loss envelope is required, evaluating the full dense grid is unnecessary.
Therefore, a surrogate fitted on observed data that approximates the loss range and sub-optimal hyperparameter regions can enable a more sparse data collection. 
We thus view the scaling law fitting procedure and its data collection as a joint hyperparameter optimization task, and propose a formulation and framework for HPO as data collection for efficient scaling law construction. 
Concretely, 

\begin{enumerate}
    \setlength{\itemsep}{1pt} 
    \setlength{\parsep}{0pt}
    \item We formalize scaling law construction as a Bayesian Optimization (BO) problem, and show that \textit{compute slicing} bridges BO with practical scaling law data collection by progressively expanding the compute window available in the acquisition process (\Cref{sec:framework}).
    \item We show that the scaling law fit can be constructed on a simulated grid, where unevaluated points are fantasized by a surrogate, removing the need to densely measure the loss envelope at every compute slice (Section \ref{sec:empirical}).
    \item We propose the metrics necessary to evaluate the search efficiency for this task, and empirically show that with the proposed changes above to a BO algorithm, it can recover the dense grid fit at a $10\text{--}100\times$ less cost than the exhaustive grid (\Cref{sec:empirical}).
\end{enumerate}
{\let\thefootnote\relax\footnotetext{Code: \url{https://github.com/OpenEuroLLM/amortizing-scaling-law-fitting-costs}}}


\section{Related Work}\label{sec:related}

Scaling laws have been a central feature of current progress in large-scale deep learning~\citep{kaplan-arxiv20a,hoffmann-neurips22a,dehghani-arxiv23a,porian-neurips24a}.
New contributions must retain their promised benefits across the scaling regime, which often necessitates either verification of existing scaling laws or construction of new fits.
Since the cost of scaling grows prohibitively, this cost is often reduced by fitting specialized parametric forms~\citep{li-neurips24a,bergsma2026power,li-arxiv25a,zhang-iclr25a}, to predict large-scale runs more accurately and thereby save compute.
Similar predictive approaches have been developed for hyperparameters and other pipeline settings~\citep{yang-neurips21a,pmlr-v235-everett24a}.
The common objective is to direct compute resources carefully while still preserving decision quality.

Hyperparameter optimization, particularly Bayesian Optimization, aims precisely at this performance-cost trade-off~\citep{frazier-arxiv18a}.
A BO algorithm has two major components: 
\begin{inparaenum}[i)]
    \item the choice of a probabilistic surrogate model (Gaussian Processes~\citep{rasmussen-book06a}, Random Forests~\citep{hutter-lion11a}, etc.), and
    \item the acquisition strategy that utilizes the surrogate prediction and uncertainty to choose the next query point.
\end{inparaenum} 
Scaling law construction can similarly be viewed as a relatively low-dimensional regression problem built on observations obtained from multiple costly deep-learning runs, with an outer procedure fitting/updating the scaling law parameters~\citep{li-iclr25a}.
Closest to our setting,~\citet{li-colm26a} formulate the selection of these runs as a budget-aware sequential experimental-design problem and propose a specialized acquisition strategy tailored for extrapolation. 
They show that observations over the modeled input space can be selected sparsely while maintaining accurate extrapolation. 
In contrast, we also consider sparsity over auxiliary hyperparameters omitted from the final scaling law form, whose optimization is required to construct the near-optimal loss frontier.
\citet{li-colm26a} consider general parametric scaling law forms, whereas we focus specifically on compute-optimal scaling laws, where each response represents the near-optimal loss obtained after optimizing training hyperparameters at a given compute scale.
Concurrently,~\citet{chen-arxiv26a} design a custom BO acquisition for scaling law construction, but only for optimal hyperparameters, addressing a subset of our general case we describe below.





\section{Framework}\label{sec:framework}

Scaling law formulations model loss $L$ as a power law in the independent variables determining compute (Appendix~\ref{app:parametric_forms}). 
Let \(N\in\mathcal{N}\), \(D\in\mathcal{D}\), and \(\lambda\in\Lambda\) denote the model size, token budget, and training hyperparameters (e.g., learning rate or batch size), respectively, and let \(C=g(N,D)\)\footnote{We adopt the commonly used approximation $C=6ND$ introduced in \cite{kaplan-arxiv20a}.} denote compute. 
For a compute-optimal scaling law fit to be predictive over large extrapolation ranges, the observed data must approximate the optimal loss $ \min{} L(C)$ at each compute scale. 
We frame the construction of these observations as a hyperparameter optimization problem (HPO), and use an iterative loop to make controlled progress in collecting data for a scaling law parametric fit.

Compared to the usual dense evaluation of the grid $\mathcal{G}=\mathcal{N}\times\mathcal{D}\times\Lambda$, we note two clear sources of efficiency gains: 
\begin{inparaenum}[(i)]
    \item avoiding evaluations of low-performing hyperparameter configurations at each compute budget $C$, parameter count $N$, or token budget $D$;
    \item approximating a dense optimal loss envelope to obtain early signal on scaling law fits, instead of evaluating the complete $\mathcal{G}$.
\end{inparaenum}

For the first source, we restrict the scope of data acquisition at each iteration, a deliberately myopic approach we refer to as \textit{compute slicing}. 
Let \(\mathcal{W}_j=[C_1,C_j]\) denote the compute window observed up to compute scale \(C_j\). 
At each iteration, the surrogate is fitted on all observations collected within \(\mathcal{W}_j\), while the search for the next configuration is performed over the expanded window \(\mathcal{W}_{j+1}\), which includes the immediate next compute slice. 
The size of a compute slice is determined by the discrete choices of \(N\) and \(D\) made for the scaling studies and is therefore fixed by the study design rather than tuned. 
We hypothesize that this allows the algorithm to conservatively invest in higher-compute configurations while retaining an opportunity to improve upon the loss observed at lower-compute slices. 
Although we do not explicitly isolate this effect within the scope of this work, the restriction is compatible with any general choice of BO components and serves only as a constraint over acquisition optimization.

For the second source, we propose an evaluation procedure at each iteration for tracking the scaling law fit improvement based on the information currently available. 
Let \(\mathcal{O}_t\) denote the observations collected by iteration \(t\), and let \(\mathcal{M}_t\) denote the surrogate fitted to them. 
The surrogate can be used to \textit{fantasize} possible outputs at unobserved parts of \(\mathcal{G}\), allowing the scaling law to be fitted on a proxy dense grid containing observed losses where available and surrogate predictions elsewhere. 
As \(\mathcal{M}_t\) improves with additional observations and better approximates the loss scales of unseen configurations, the resulting fit can approach the full data fit at a fraction of the total cost.

Therefore, a well-designed surrogate that approximates the scale and shape of the loss landscape can enable efficient scaling law data collection.
Our subsequent section shows empirical evidence that clear efficiency gains are achievable by modifying vanilla BO with two core mechanisms: compute slicing and fantasized fits. We provide the complete formalization in Appendix~\ref{app:framework} and illustrate the framework in Figure~\ref{fig:framework-flowchart}.

\section{Empirical Analysis}\label{sec:empirical}



\begin{figure}[htbp]
    \centering
    \includegraphics[width=\linewidth]{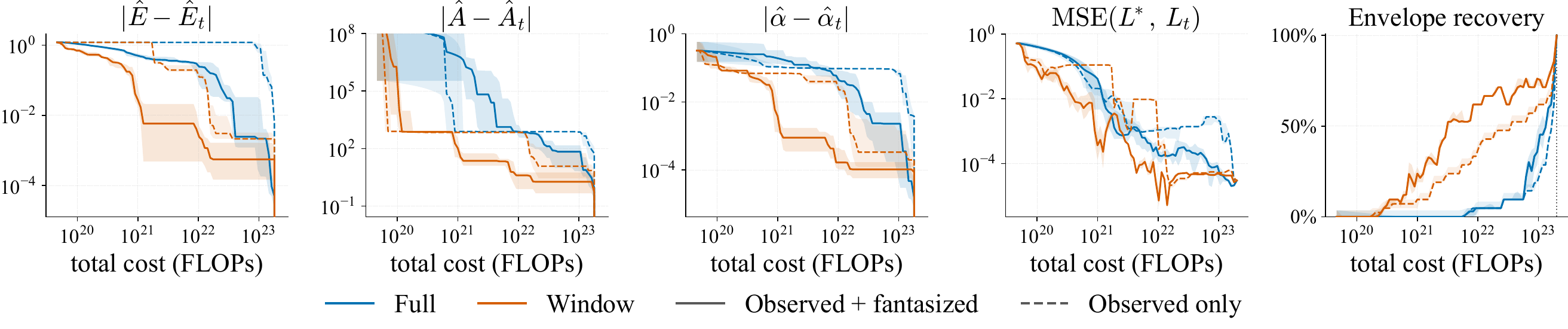}
    \vspace{-20pt}
    \caption{
    Results on $\chinOne$ comparing the same BO surrogate model, acquiring data by optimizing over the \textit{Full} search space or a growing \textit{Window} conditioned on the observations so far.
    Panels 1-3 report absolute error in each fitted parameter against the full data fit (sans \textit{held-out}). 
    Panel 4 records the MSE of the loss predicted by the fitted parameters against the actual held-out loss.
    Panel 5 tracks the \% recovery of the optimal-loss envelope.
    Shaded band is 1 S.D. over 10 seeds.
    }
    \label{fig:main_result}
\end{figure}

\paragraph{Setup.} We rely on existing collections of scaling experiments and simulate scaling law construction over their predefined configuration grids. Our goal is to assess whether the proposed framework enables a standard search algorithm to recover scaling law fits comparable to those obtained from the full grid while using substantially less cumulative compute. 
We evaluate 4 settings: compute \textit{Window}, the standard BO definition of viewing the entire search space \textit{Full}, and with or without fantasization of pending configurations.
These configurations are compared in terms of scaling law recovery as a function of cumulative compute. 
In~\Cref{fig:main_result}, we present results on OELLM-English~\cite{ajroldi-arxiv26a} using a standard BoTorch~\citep{balandat-neurips20a} defaults (Gaussian Process surrogate with constant mean, Mat\'ern-5/2 kernel, and a Lower Confidence Bound acquisition with $\kappa=2$). We repeat the study across additional datasets, parametric forms, and search methods detailed in Appendix~\ref{app:exp} and Appendix~\ref{app:ablation}.

\paragraph{Metrics.} Just as in standard HPO, our framework seeks to minimize observed loss, albeit for each compute slice, while additionally filling in the dense grid through fantasization and evaluating the resulting scaling law fit at each iteration. 
Existing literature typically reports the in-sample residual fit or the extrapolation error on the held-out set alone~\citep{li-colm26a}. 
We argue that neither is sufficient on its own, since improvements in one do not translate predictably into the other, as they are measured on different scales. 
Depending on the parametric form and data, different fits can yield similar errors on the immediate held-out set but vary more drastically at much larger compute horizons. 
Since the analysis samples from a fixed grid, every search method eventually observes the entire data and all evaluation metrics converge once the entire grid is exhausted, so only the trajectory over cumulative compute separates methods.
We present empirical evidence for why all metrics need to be studied \textit{jointly} in the next section and use the following:
\begin{inparaenum}[(i)]
    \item Absolute difference of the estimated scaling law parameters to the ground truth estimated parameters
    (Regret);
    \item The mean-squared-error of the prediction on the held-out points of a fitted law at an iteration to the empirical loss recorded in the dataset (MSE);
    \item The percentage coverage of the ground truth loss-envelope set; 
    \item Relative \% difference in loss prediction over large extrapolation ranges.
\end{inparaenum}
We detail our metrics in Appendix~\ref{app:exp-metrics}.

\paragraph{Results.} 
To assess the benefits of our proposed framework employing a growing compute window (\textit{Window}), we use the standard BO definition of viewing the entire space of $(N, D, \Lambda)$ as a uniform search space (\textit{Full}).
Figure~\ref{fig:main_result} shows that across both strategies and both fit types, the coefficients are recovered well before the pool is exhausted, and a small fraction of the total acquisition compute already yields coefficient estimates close to the full data reference.
The speedup for \textit{Window+fantasization} is $10\times$ when compared to \textit{Full+fantasization}, and up to even $100\times$ when compared to \textit{Full+observed-only}.
The envelope coverage panel indicates that this efficiency comes from where we expect it, with
\textit{window} discovering its evaluations on the loss envelope far earlier than \textit{full},
which is the behaviour compute slicing is designed to produce.

Table~\ref{tab:extrap-compute-lcb} presents whether these recovered coefficients still agree once extrapolated well beyond the predefined grid, where the budget is reported as a fraction of the compute required to exhaust the grid.
At $1\%$ of total budget, the predicted losses stay within $1.6\%$ of the full data fit even at $6$--$7$ orders of magnitude higher.
Spending more budget provides only marginal relative  (coefficients in Table~\ref{tab:lcb-incumbent-coeffs}).

\begin{table}[pbth]
\centering
\caption{
Extending the results of~\Cref{fig:main_result}'s \textit{window + fantasization} to much larger extrapolation lengths.
Entries are the relative \% difference in predicted loss against the full data fit (sans held-out), at different fractions of the compute to exhaust the grid.
S.D. over 10 seeds.
}
\label{tab:extrap-compute-lcb}
\begin{tabular}{lccc}
\toprule
Compute $C$ (in FLOPs): & $10^{25}$ & $10^{27}$ & $10^{29}$ \\
\midrule
$1\%$ of Total Budget  & $1.09 \pm 1.84$ & $1.37 \pm 2.35$ & $1.56 \pm 2.67$ \\
$5\%$ of Total Budget & $0.22 \pm 0.34$ & $0.29 \pm 0.41$ & $0.36 \pm 0.45$ \\
$10\%$ of Total Budget  & $0.04 \pm 0.04$ & $0.05 \pm 0.04$ & $0.06 \pm 0.10$ \\
\bottomrule
\end{tabular}
\end{table}



\section{Conclusion \& Future Work}\label{sec:conclusion}

We cast scaling law construction as a hyperparameter optimization (HPO) problem.
We demonstrate that progressively growing the compute window during acquisition, combined with surrogate fantasization to evaluate scaling law fits on a simulated dense grid proxy, allows 
BO 
to achieve highly accurate fits while reducing compute costs by $10$--$100\times$ over the actual exhaustive grid. 
The recovered parameters' predicted losses remain within $1.6\%$ of the full-data fit, even at $6$--$7$ orders of magnitude beyond the observed grid.
To properly evaluate this efficiency, we introduced a multi-metric suite that captures the nuances of scaling law fits.
Our ablations confirm that our proposed framework performs well for different settings and parametric forms, and seamlessly integrates with standard BO formulations.
While the development of specialized kernels and acquisition functions for this framework presents an exciting direction 
to explore,
we treat this as future work, and the ultimate stress-test will be validating this design on a novel, large-scale data collection run for scaling law discovery.
An acquisition function that additionally incorporates scaling law fit variance is a natural extension that we 
also
leave to future work.





\section*{Acknowledgements}

We thank Johannes Hog for their feedback on the paper draft.

\textbf{NM} acknowledge funding by 
the state of Baden-W\"{u}rttemberg through bwHPC, the German Research Foundation (DFG) through grant numbers INST 39/963-1 FUGG and 417962828, and the European Union (via ERC Consolidator Grant Deep Learning 2.0, grant no.~101045765). 
Views and opinions expressed are however those of the author(s) only and do not necessarily reflect those of the European Union or the European Research Council. 
Neither the European Union nor the granting authority can be held responsible for them.
This research was also funded by the Deutsche Forschungsgemeinschaft (DFG, German Research Foundation) under grant number 539134284, through EFRE (FEIH\_2698644) and the state of Baden-Württemberg.
\textbf{NM} is supported by the Konrad Zuse School of Excellence in Learning and Intelligent Systems (ELIZA) through the DAAD programme Konrad Zuse Schools of Excellence in Artificial Intelligence, sponsored by the Federal Ministry of Education and Research.
\textbf{AK, JV, AJ, DO, SH, SL, NA} acknowledge support from EC under the grant No. 101195233 (OpenEuroLLM). \textbf{DO, JV} acknowledge that this publication is part of project 2026.008 of the Computing Time National Computing Facilities program, which is financed by the Dutch Research Council (NWO) under the grant \href{https://doi.org/10.61686/WBFDE48237}{https://doi.org/10.61686/WBFDE48237}. 
\begin{center}
\includegraphics[width=0.25\textwidth]{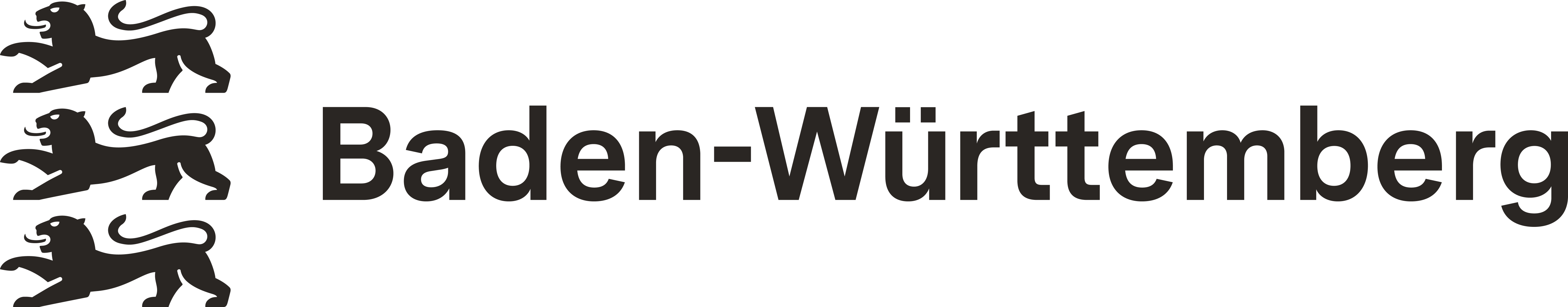} \quad
\includegraphics[width=0.3\textwidth]{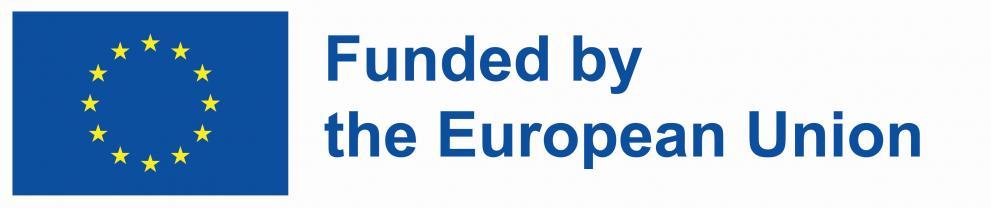} \quad
\includegraphics[width=0.12\textwidth]{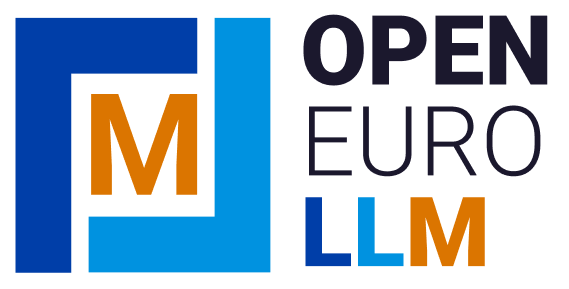}
\end{center}


\newpage

\bibliography{bib/strings,bib/lib,bib/local,bib/proc}
\bibliographystyle{abbrvnat}


\newpage
\appendix




\section{Limitations of the Current Study}\label{app:limitations}


There are several limitations in our current manuscript that can lead to many research directions for future extensions. 
First, we acknowledge that our results are a replay of existing dense grids, not a live data collection strategy for discovering a new law at a previously unmeasured scale for new modeling or data interventions. 
Some of the metrics we track (e.g., parameter differences) also lose meaning without a reference fit. 
We simulate different scenarios via ablations on other scaling datasets in Appendix~\ref{app:exp-datasets}.

Second, methodologically, we operate with standard kernel and acquisition function
choices for Bayesian optimization, and offer some simple ablations in Appendix~\ref{app:kernel choice} and Appendix~\ref{app:acquisition_function}.
There are potential extensions to scaling law form-aware mean and kernel functions (e.g., the group additive kernel used in Section~\ref{app:kernel choice}) for the GP. 
We also do not fully exploit the uncertainty measures available from the surrogate in scaling law construction. 
Potential extensions could propagate the uncertainty from the fantasized grid to scaling law coefficients directly, or weigh them differently in the weighted regression problem for identifying the fit.

Third, methods to construct scaling laws by discarding a large part of a dense grid can still offer insights on how better to design scaling law studies. 
The choice between trying out a new hyperparameter setting, extending a training run longer, and moving to a new model size dictates how we measure the best loss at a given compute budget. 
This directly ties to the width of the window used during data collection in this work. 
We also note that our compute \textit{slicing} setup is similar in principle to a cost-aware acquisition.
However, when measuring cost in FLOPs, multiple hyperparameters may share similar costs and therefore may not suitably change ranking in the acquisition function.




\section{Framework (continued)}\label{app:framework}


\paragraph{Notation Summary} We provide all the notations used within this study in~\Cref{tab:notation}.

Typical scaling law studies define a scaling ladder, that is, a grid of model sizes and token budgets, contingent on the architecture chosen, hardware used, and total budget available.
Let $\mathcal{N} = \{N_1, N_2, \ldots, N_n\} \subset \mathcal{Z}_+$ and $\mathcal{D} = \{D_1, D_2, \ldots, D_d\} \subset \mathcal{Z}_+$ define the space of model and token budgets, respectively, and $\Lambda \subseteq \mathbb{R}^k$ defines a space of hyperparameters considered.
Let $f: \mathcal{N} \times \mathcal{D} \times \Lambda \rightarrow \mathbb{R}$ represent the loss of a model run.
A standard HPO problem would then be,

\begin{equation}\label{eq:hpo-vanilla}
(N, D, \lambda)^{*} \in
\argmin_{N \in \mathcal{N},\, D \in \mathcal{D},\, \lambda \in \Lambda}
f(N, D, \lambda).
\end{equation}

Given that compute is some function of model parameters and data seen, let $C(N, D) = g(N, D) \in \mathbb{R}_{+}$
\footnote{e.g. $g(N, D) = 6 \cdot N \cdot D$ from~\citet{kaplan-arxiv20a}}
, then the~\Cref{eq:hpo-vanilla} for scaling law data optimization will be,

\begin{equation}\label{eq:hpo-scaling}
(N, D, \lambda)^{*}_{\mathcal{S}_j} \in
\argmin_{\substack{N \in \mathcal{N},\, D \in \mathcal{D},\, \lambda \in \Lambda \\ g(N, D) \in \mathcal{S}_j}}
f(N, D, \lambda),
\qquad \forall j = 1, \dots, m.
\end{equation}

Let $g: \mathcal{N} \times \mathcal{D} \to \mathbb{R}_{+}$ map a configuration to its compute, and let $C_1 < C_2 < \ldots < C_m$ be the distinct compute levels it induces, forming an increasing compute frontier (typically iso-FLOP, and generalizable to iso-parameter, iso-token, etc.).
Since $g$ is many-to-one, several $(N, D)$ pairs may share a compute level, so $m \leq n \cdot d$, with equality only when no two grid points collide.
Let a \textit{compute window} be defined as $\mathcal{W}_{j} = [\,C_{1},\, C_{j}\,]$, so that $\mathcal{W}_1 \subset \mathcal{W}_2 \subset \ldots \subset \mathcal{W}_{j} \subset \ldots \subset \mathcal{W}_m$.
Each consecutive pair of windows defines a compute \textit{slice}, $\mathcal{S}_j = \mathcal{W}_{j} \setminus \mathcal{W}_{j-1} = (\,C_{j-1},\, C_{j}\,]$ for $j > 1$, with $\mathcal{S}_1 = \mathcal{W}_1$.

The full space of optimization is thus the grid $\mathcal{G} = \mathcal{N} \times \mathcal{D} \times \Lambda$.
We label any HPO that searches this space jointly as \texttt{Full} in our analysis.
Let $\mathcal{O}_t$ be the set of observations collected during an HPO run up to iteration $t$.
For the compute window-based HPO, the search happens over $\mathcal{W}_{p+1}$, where $p = \max \{\, j : \mathcal{S}_j \cap \mathcal{O}_t \neq \emptyset \,\}$ is the index of the highest compute slice observed so far.
In our analysis, we call such methods \texttt{Window}.

For evaluation, given a compute threshold $\tau$ as input, the full optimization space is split into a pool $\mathcal{G}_{\mathrm{pool}} = \{\, x \in \mathcal{G} : C(x) < \tau \,\}$ and a held-out set $\mathcal{G}_{\mathrm{held\text{-}out}} = \{\, x \in \mathcal{G} : C(x) \geq \tau \,\}$.
The search is restricted to the pool, so $\mathcal{O}_t \subseteq \mathcal{G}_{\mathrm{pool}}$ at every iteration.
When a standard scaling law fit procedure is executed on $\mathcal{O}_t$ alone, we call it \texttt{Observed}.

For model-based HPO, a surrogate $\mathcal{M}_t$ is fitted on $\mathcal{O}_t$ and used to \textit{fantasize} predictions for the unevaluated points $\mathcal{G}_{\mathrm{pool}} \setminus \mathcal{O}_t$.
Executing the scaling law fit procedure on this simulated pool, which takes observed losses where available and surrogate predictions elsewhere, gives the fit we call \texttt{Observed+Fantasized}.


\begin{table}[htbp]
\centering
\caption{Summary of notations used.}
\label{tab:notation}
\begin{tabular}{ll}
\toprule
Symbol & Definition \\
\midrule
$N$, $D$ & Model size (parameter count) and training tokens \\
$\lambda$ & Hyperparameter configuration, $\lambda \in \Lambda \subseteq \mathbb{R}^k$ \\
$\mathcal{N}$, $\mathcal{D}$ & Sets of candidate model sizes and token budgets \\
$\mathcal{G}$ & Full grid of configurations, $\mathcal{G} = \mathcal{N} \times \mathcal{D} \times \Lambda$ \\
$f(N, D, \lambda)$ & Loss of a training run \\
$g(N, D)$ & Compute cost in FLOPs, $C(N,D) = g(N,D) \approx 6ND$ \\
$C_1 < \ldots < C_m$ & Distinct compute levels induced by $g$ over the grid \\
$\mathcal{W}_j$ & Compute window, $\mathcal{W}_j = [C_1, C_j]$ \\
$\mathcal{S}_j$ & Compute slice, $\mathcal{S}_j = \mathcal{W}_j \setminus \mathcal{W}_{j-1}$ \\
$\tau$ & Compute threshold separating the pool from the held-out set \\
$\mathcal{G}_{\mathrm{pool}}$ & Configurations available to the search, $\{x \in \mathcal{G} : C(x) < \tau\}$ \\
$\mathcal{G}_{\mathrm{held\text{-}out}}$ & Configurations reserved for extrapolation testing, $\{x \in \mathcal{G} : C(x) \geq \tau\}$ \\
$\mathcal{O}_t$ & Observations collected up to iteration $t$, $\mathcal{O}_t \subseteq \mathcal{G}_{\mathrm{pool}}$ \\
$\mathcal{M}_t$ & Surrogate model fitted on $\mathcal{O}_t$ \\
\bottomrule
\end{tabular}
\end{table}

\begin{figure}[htbp]
    \centering
    \includegraphics[width=0.9\linewidth]{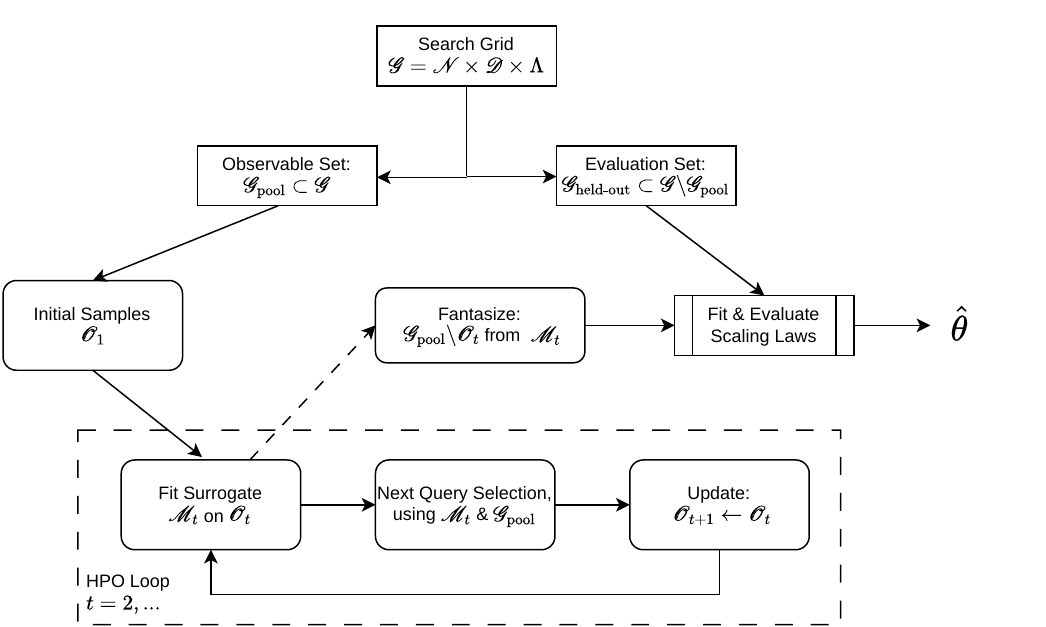}
    \caption{Overview of the framework. From the search grid we draw initial samples, then run the
HPO loop: at each iteration a surrogate $M_t$ is fit on the observations $O_t$, the next query is
selected over the expanding window $W_{t+1}$, its loss is evaluated, and $O_t$ is updated, until
the compute budget $C_{\mathrm{total}}$ is reached. The surrogate optionally fantasizes losses
over the grid, and the resulting observed (and fantasized) points are used to fit and evaluate
the scaling law against the full-grid reference. Compute slicing and fantasization are the two
additions to vanilla BO.}
    \label{fig:framework-flowchart}
\end{figure}

Figure~\ref{fig:framework-flowchart} assembles these components into the full loop: starting from an
initial design within the first compute window, each iteration fits the surrogate $M_t$ on the
observations $O_t$, selects and evaluates the next configuration over the expanded window $W_{t+1}$, and optionally (in case of model-based method) fantasizes the unobserved pool $G_{\mathrm{pool}}\setminus O_t$ to fit and score the scaling law at that step.
The loop repeats until the compute budget is spent, after which the recovered parameters $\hat\theta$ are evaluated against the full-grid reference, with compute slicing (the window $W_{p+1}$) and fantasization ($M_t$) as the only two additions to an otherwise standard BO procedure.

\section{Experimental Extras}\label{app:exp}
To assess the robustness of our framework, we validate the proposed method on several dimensions, namely different datasets (Appendix~\ref{app:exp-datasets}) and  parametric forms (Appendix \ref{app:parametric_forms}). Additionally, we detail the metrics used to track within our experiments (Appendix \ref{app:exp-metrics}) and the experiment settings used for motivating our analysis (Appendix \ref{app:experimental_settings}). 

\subsection{Datasets Used}\label{app:exp-datasets}

We evaluate the proposed framework on 5 datasets spanning different compute ranges, grid densities, and scaling study designs. These differences are important because they determine which scaling laws can be reliably recovered from each dataset and affect the difficulty of the data acquisition problem.

\paragraph{OELLM-English} \cite{ajroldi-arxiv26a} OELLM-English provides nearly complete coverage of the $(N, D)$ grid, with 51 of 54 possible combinations (94\%) represented in Table \ref{tab:datasets_overview}. It also features a convenient scaling geometry for the compute slicing: token budgets are approximately geometrically spaced, with successive budgets differing by roughly $1.5 - 1.7 \times$, resulting in approximately uniform spacing along the log-compute axis. Its full sweep has one of the largest total collection costs among the considered datasets, comparable to StepLaw (Table \ref{tab:datasets_scales}). The dataset was designed to study loss scaling with compute, model size and token budget, as well as the scaling of optimal learning rate and batch size scaling across $(N, D)$.

\paragraph{Porian-HP} \cite{porian-neurips24a} From the experiments of \cite{porian-neurips24a}, we select the hyperparameter scaling dataset which evaluates learning rate, batch size and Adam $\beta_2$ across multiple model and data scales. Porian-HP requires substantially lower compute budgets than the other datasets and provides a sparse coverage of the $(N, D)$ space, containing only 14\% of the possible combinations. Moreover, token budget is not varied independently of model size: runs follow similar token-per-ratios; thus, the observations cover a narrow region of the $(N, D)$ grid. This limited independent variation in $N$ and $D$ makes the dataset unsuitable for fitting scaling laws that require two-dimensional $(N, D)$ grid coverage, such as Chinchilla Approach 3 \cite{hoffmann-neurips22a}. The 566 runs additionally vary with Adam $\beta_2$, resulting in roughly 237 distinct $(N, D, \eta, b)$ configurations. The dataset was specifically designed to study the scaling of optimal hyperparameters.

\paragraph{StepLaw}~\citep{li-arxiv25a} StepLaw provides a broad coverage of learning rate and batch size, consisting of 175 of 182 possible $(\eta, b)$ combinations (96\%), seen in Table \ref{tab:datasets_overview}. In contrast, its $(N, D)$ coverage is sparse: only 23\% of the possible combinations are represented, spanning 5 model sizes over a relatively narrow range. Nevertheless, the large number of hyperparameter evaluations at these $(N, D)$ points makes StepLaw particularly well suited for evaluating hyperparameter search methods. Its full sweep has one of the highest total compute cost in our collection, comparable to OELLM-English (\ref{tab:datasets_scales}). The dataset was collected to study the scaling of optimal hyperparameters.

\paragraph{Warmstart}~\citep{mallik-neurips26a} 
Its $(N, D)$ coverage is sparse, with only 17\% of possible combinations represented, and only $23$ distinct $(N,D)$ pairs.
Unlike other datasets, here $\mu$P~\citep{yang-neurips21a} and batch size scaling~\cite{zhang-iclr25a} was used so a typically dense $(\eta,~b)$ grid is not available.
This project collected runs on language models that were initialized from smaller trained language models.
Therefore, this dataset has $4$ other non-standard hyperparameters: the size of the smaller model used to initialize, the method to warmstart, warmstart hyperparameters, and the ratio of hidden sizes of the small and large model. The dataset was collected to study the scaling of loss with respect to model size and the token budget.

\paragraph{OELLM-Multilingual} \cite{oellm_multilingual_scaling} OELLM-Multilingual follows a similar experimental design to OELLM-English, varying model size, token budget, learning rate and batch size. It provides complete coverage of its $(N,D)$ grid, with all 36 possible combinations represented. Compared with OELLM-English, however, it spans fewer model sizes and evaluates a smaller hyperparameter sweep around each $(N,D)$ pair. It therefore retains regular two-dimensional model-data scaling coverage while providing a less densely sampled hyperparameter search problem. The dataset was designed to study loss scaling with compute, model size and token budget, together with the scaling of optimal hyperparameters in the multilingual setting.

\noindent\textit{Note.} Tuple counts in Table~\ref{tab:datasets_overview} denote the number of distinct combinations observed in each dataset and therefore need not equal the product of the corresponding marginal counts. Differences between the total number of runs and the number of distinct $(N,D,\eta,b)$ configurations arise from additional variables varied in the original experimental design. For example, Porian-HP additionally sweeps three values of Adam $\beta_2$.

\begin{table}[h!]
\centering
\caption{Overview of the benchmark datasets and their configuration grids.}
\begin{tabular}{lcccccccc}
\toprule
\textbf{Dataset} & \textbf{\# Runs} &
\textbf{\# $N$} & \textbf{\# $D$} &
\textbf{\# $(N,D)$} & \textbf{\# $\eta$} & \textbf{\# $b$} &
\textbf{\# $(\eta,b)$} & \textbf{\# $(N,D,\eta,b)$} \\
\midrule
OELLM-English     & 909  & 6 & 9  & 51 & 5  & 8  & 36  & 942  \\
Porian-HP   & 566  & 7 & 39 & 39 & 7  & 7  & 46  & 237  \\
StepLaw      & 1911 & 5 & 15 & 17 & 14 & 13 & 175 & 1911 \\
Warmstart  & 150  & 6 & 23 & 23 & 3  & 7  & 14  & 44   \\
OELLM-Multilingual & 349  & 4 & 9  & 36 & 5  & 8  & 30  & 332  \\
\bottomrule
\end{tabular}
\label{tab:datasets_overview}
\end{table}

\begin{table}[h!]
\centering
\caption{Scale and compute coverage of the benchmark datasets.}
\begin{tabular}{lccc}
\toprule
\textbf{Dataset}
& \textbf{$N$ range}
& \textbf{$D$ range}
& $C_{\mathrm{total}}$ \\
\midrule

OELLM-English
& $47.6\mathrm{M}$--$1.71\mathrm{B}$
& $6\mathrm{B}$--$300\mathrm{B}$
& $1.45 \times 10^{23}$ \\

Porian-HP
& $5.2\mathrm{M}$--$221\mathrm{M}$
& $103\mathrm{M}$--$4.4\mathrm{B}$
& $2.88 \times 10^{20}$ \\

StepLaw
& $215\mathrm{M}$--$1.07\mathrm{B}$
& $4\mathrm{B}$--$100\mathrm{B}$
& $1.45 \times 10^{23}$ \\

Warmstart
& $32.3\mathrm{M}$--$1.22\mathrm{B}$
& $32.3\mathrm{M}$--$60.9\mathrm{B}$
& $1.66 \times 10^{21}$ \\

OELLM-Multilingual
& $206\mathrm{M}$--$1.26\mathrm{B}$
& $6\mathrm{B}$--$300\mathrm{B}$
& $6.57 \times 10^{22}$ \\
\bottomrule
\end{tabular}
\label{tab:datasets_scales}
\end{table}

\subsection{Parametric Forms}
\label{app:parametric_forms}

We evaluate scaling law construction using two parametric forms that model loss at
increasing scale. The forms differ in how the scale variables are represented
and, consequently, in which configurations from the experimental grid contribute
to the fit. For variables that are not explicitly modeled by the parametric form,
we select the configuration achieving the lowest loss.

\subsubsection{Loss as a function of compute: $L(C)$}

We first consider a power law relating loss directly to training compute,
\begin{equation}
    L(C) = E + A C^{\alpha},
\end{equation}
where $E$ denotes the irreducible loss, $A$ controls the scale of the
compute-dependent term, and $\alpha < 0$ determines the scaling rate.

Since this formulation models loss only as a function of compute, configurations
that differ in model size, token budget, or training hyperparameters but have
the same compute are represented by the lowest-loss configuration. For each
compute level $C$, we therefore define
\begin{equation}
    L^\star(C)
    =
    \min_{x \in \mathcal{G}: C(x)=C} L(x).
\end{equation}
The scaling law is fitted to the resulting best-loss envelope
$\{(C,L^\star(C))\}$. Thus, $L(C)$ collapses the allocation of compute between
model size and training tokens, as well as the training hyperparameters, into a
one-dimensional compute-loss relationship.

\subsubsection{Loss as a function of model size and data: $L(N,D)$}

We additionally evaluate the parametric loss model proposed in Approach 3 of
Chinchilla~\cite{hoffmann-neurips22a},
\begin{equation}
    \chinThree
\end{equation}

%
The five parameters $(E,A,B,\alpha,\beta)$ separately characterize the
dependence of loss on model size $N$ and training tokens $D$.

Unlike $L(C)$, this formulation retains the two-dimensional $(N,D)$ structure
of the scaling study. Since our experimental grids additionally contain
training hyperparameters, we first select the lowest-loss configuration for
each $(N,D)$ pair,
\begin{equation}
    L^\star(N,D)
    =
    \min_{\eta,b} L(N,D,\eta,b),
\end{equation}
and fit the parametric form to the resulting set
$\{(N,D,L^\star(N,D))\}$.

Under the approximation $C \approx 6ND$, the fitted parameters additionally
determine the compute-optimal allocation between model size and training data,

Consequently, recovering $L(N,D)$ places a stronger requirement on the
coverage of the experimental grid than recovering $L(C)$: the acquisition
process must recover near-optimal losses across a sufficiently broad set of
$(N,D)$ pairs rather than only the one-dimensional compute-loss envelope.

This formulation estimates 5 parameters, and models the dependence of loss on model size $N$ and training tokens $D$ separately. 

This setting differs from the $L(C)$ scaling law considered above. While $L(C)$ is fitted to the best-loss frontier across compute scales, $L(N, D$) retains the 2D $(N, D)$ structure. Since our search space additionally contains training hyperparameters, we therefore identify the best performing configuration for each $(N, D)$ pairs before fitting the scaling law. 







\subsection{Metrics}\label{app:exp-metrics}

\paragraph{Why look at multiple metrics?}
Progress in HPO is typically measured by the minimum loss achieved, matching its underlying $\arg\min L$ objective.
We argue this is not enough for scaling law construction:
An acquisition strategy can drive the loss down by clustering samples at the highest compute scale, leaving lower compute scales sparse 
which can affect the parametric fit quality.
We therefore keep the usual check of what loss the acquired points actually achieve, but also treat it as one measure among several.
Our primary metrics thus do not track traditional loss regret, instead they measure discrepancies (e.g., parameter error, frontier recovery, and held-out predictive loss) relative to a reference fit as defined below, 
extending notations from~\Cref{tab:notation}.

\paragraph{Set up.} 
To extract the loss envelope on any grid $\mathcal{G}'$, we define the frontier operator $\Pi(\mathcal{G}')$: to fit $L(C)$ forms, $\Pi(\mathcal{G}')$ extracts the compute--loss Pareto frontier; for $L(N,D)$ forms, $\Pi(\mathcal{G}')$ extracts the min-loss configuration per $(N,D)$ cell present in $\mathcal{G}'$.
More generally, $\Pi(\mathcal{G}')$ extracts the min-loss over all dimensions in $\mathcal{G}'$ that is not an independent variable modeled in the loss-parametric form.

\paragraph{Reference fits.}
Given our empirical analysis has the full grid evaluation available, we compute two oracle fits on available reference grids: the \textit{validation} fit $\hat{\theta}_{\mathrm{val}}$ fitted on $\Pi(\mathcal{G}_{\mathrm{pool}})$, and the \textit{global} fit
$\hat{\theta}_{\mathrm{global}}$ fitted on $\Pi(\mathcal{G})$.
At each acquisition step, a surrogate $\mathcal{M}_t$ can fantasize losses across the unobserved pool to form a mixed (observed + fantasized) grid $\hat{\mathcal{G}}$. The estimated law $\hat{\theta}$ is then fitted directly on the extracted mixed frontier $\Pi(\hat{\mathcal{G}})$.
Unless otherwise stated, we omit the subscript $\mathrm{val}$ from parameters $\hat{\theta}$ in the plots.

To measure the quality of the fits, we propose to use 4 metrics.


\textbf{Parameter regret.} The unsigned parameter discrepancy $r(\hat\theta_t, \hat{\theta}_{\mathrm{val}}) = \lvert\hat\theta_t - \hat{\theta}_{\mathrm{val}}\rvert$, reported as the running incumbent.

\textbf{Held-out error.} The mean squared error between predictions from $L_{\hat{\theta}_t}$ (often abbreviated as $L_t$ in plots) and the ground-truth values $L^*$ on the held-out frontier $\Pi(\mathcal{G}_{\mathrm{held-out}})$.

\textbf{Envelope recovery.} The fraction of the true envelope recovered by the mixed grid, given by $R / |\Pi(\mathcal{G}_{\mathrm{pool}})|$ where $R = |\Pi(\hat{\mathcal{G}}) \cap \Pi(\mathcal{G}_{\mathrm{pool}})|$. (Replacing $\hat{\mathcal{G}}$ with acquired points yields the observed-only variant). This fraction measures whether the mixed (observed + fantasized) grid has recovered the points the reference fit is trained on.

\textbf{Relative Error Extrapolation.} To have a more complete picture of how the scale of regret over the parameter fits affects the extrapolation quality, we also compare the relative \% error at larger computes for the fit obtained on $\Pi(\hat{\mathcal{G}})$ to the fit obtained on $\Pi(\mathcal{G}_{\mathrm{pool}})$.

\begin{table}[pbth]
\centering
\caption{
Incumbent coefficients of the loss law $L(C)=E+A\,C^{\alpha}$ under LCB
acquisition, at increasing fractions of the compute needed to
exhaust the grid. Mean $\pm$ S.D.\ over 10 seeds.
}
\label{tab:lcb-incumbent-coeffs}
\begin{tabular}{lccc}
\toprule
Budget & $E$ & $A$ & $\alpha$ \\
\midrule
$1\%$ of Total Budget   & $0.4655 \pm 0.0309$ & $6.833 \pm 0.332$ & $-0.1559 \pm 0.0087$ \\
$5\%$ of Total Budget   & $0.4536 \pm 0.0054$ & $6.686 \pm 0.024$ & $-0.1521 \pm 0.0007$ \\
$10\%$ of Total Budget  & $0.4502 \pm 0.0014$ & $6.674 \pm 0.006$ & $-0.1517 \pm 0.0001$ \\
$25\%$ of Total Budget  & $0.4502 \pm 0.0014$ & $6.674 \pm 0.006$ & $-0.1517 \pm 0.0001$ \\
$50\%$ of Total Budget  & $0.4502 \pm 0.0014$ & $6.674 \pm 0.006$ & $-0.1517 \pm 0.0001$ \\
\midrule
$100\%$ of Total Budget & $0.4499$ & $6.679$  & $-0.1518$ \\
\bottomrule
\end{tabular}
\end{table}

\subsection{Experimental Settings}
\label{app:experimental_settings}

For both $L(C)$ and $L(N, D)$ studies we evaluate on five datasets: OELLM-English, OELLM-Multilingual, StepLaw, Porian-HP, and Warmstart. 
In every case the grid is split on compute into an acquirable pool $G_{\mathrm{pool}}$ and a held-out set $G_{\mathrm{held\text{-}out}}$, where the held-out set is the top slice of the compute axis reserved for held-out error and never acquired.
We place this boundary at half the maximum available compute; a held-out fraction of $0.5$ as a standard, untuned choice that still leaves more than one point on the held-out compute-optimal frontier, so that held-out error is measured over more
than a single point.
The one exception is Warmstart, whose grid contains a single run orders of magnitude above the rest; at a fraction of $0.5$ this leaves only that lone point in the held-out set, so we widen the fraction to $0.9$ to recover a held-out frontier with more than one point.
The growing compute window $W_{t+1}$ expands with the observed frontier; its forward reach beyond
the current frontier is set per dataset to a factor of $1.5\times$ for OELLM-English and
OELLM-Multilingual and $2\times$ for StepLaw, Porian-HP, and Warmstart, while the full variant (vanilla) searches the whole pool.
Each run starts from ten practice-initialized points and acquires under a Gaussian-process surrogate over $(N, D, \lambda)$ with a
Mat\'ern kernel and Expected Improvement (for all ablations) and Lower Confidence Bound (LCB) for Figure \ref{fig:main_result}, repeated over 10 seeds; we track coefficient regret,
held-out MSE, and envelope-set coverage against cumulative compute and read them jointly.

\section{Ablations}\label{app:ablation}

In the main paper, we have shown results on the OELLM-English dataset ~\cite{ajroldi-arxiv26a}
for Gaussian processes (GP) with Mat\'ern kernel as the surrogate and Lower Confidence Bound (LCB) as the acquisition function as the 
choices for vanilla BO components.

Now we investigate whether the proposed approach 
improves general search methods, reliably identifies
scaling law fits on other datasets (Section \ref{sec:app-datasets-base} and \ref{sec:chin-datasets}) and parametric forms (Section \ref{sec:app-chin}), and we also discuss the GP modeling choices (kernel
and acquisition functions).

\subsection{Improving any BO based search method}

\begin{figure}[t]
    \centering
    \includegraphics[width=\linewidth]{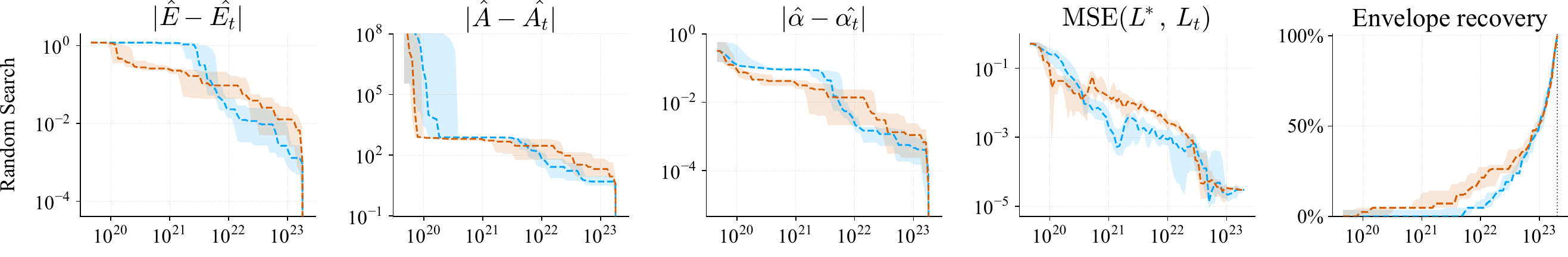}
    \includegraphics[width=\linewidth]{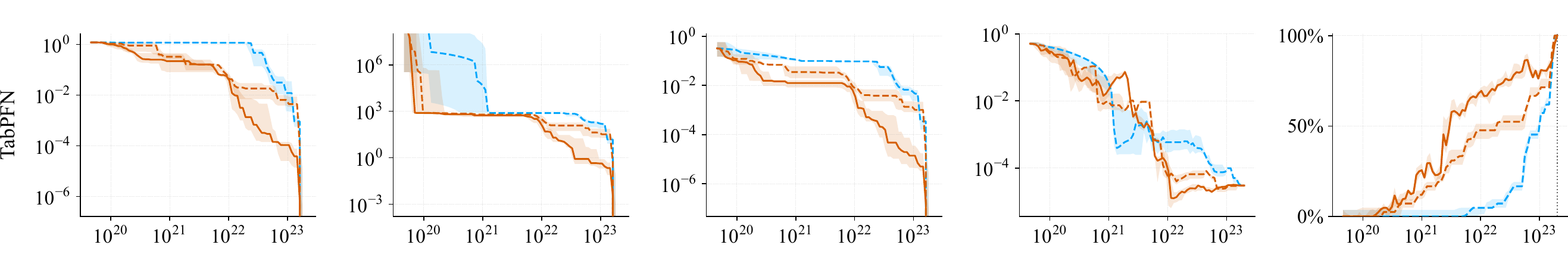}
    \includegraphics[width=\linewidth]{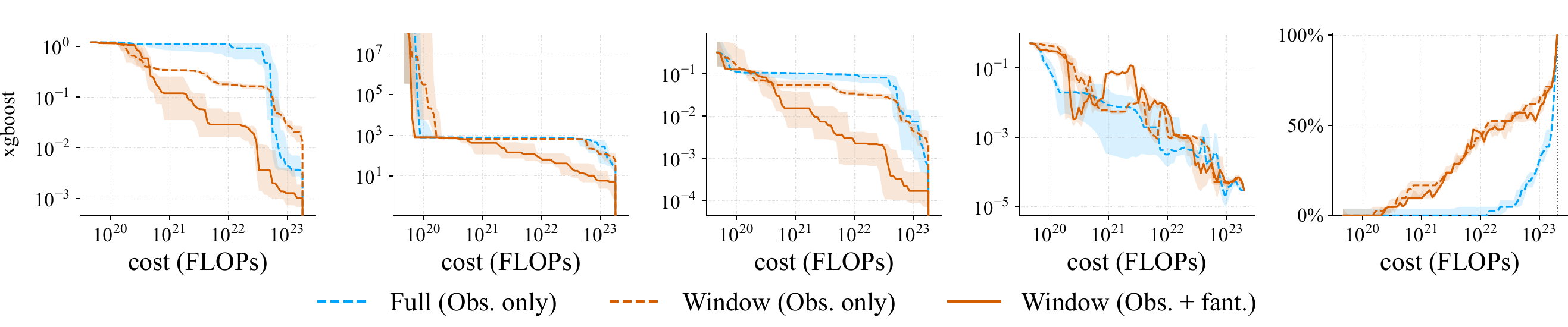}
    \caption{
    Effect of our proposed search (window) methodology with other models: Random Search (top) as a surrogate-free approach, and two other choices of surrogates for BO: TabPFN (middle) and xgboost (bottom). Our proposed method emperically improve over the standard BO definition (Full).
    }
    \label{fig:app-generalized-methods}
\end{figure}

We ask whether our proposed changes to the standard
Bayesian optimization methodology apply and improve other baseline
search procedures for efficient scaling law construction. In Figure~\ref{fig:app-generalized-methods}, we show the effect of applying compute slicing followed by fantasization (where applicable) to
Random Search and two other choices of surrogates for BO on the OELLM-English study. We consider ensemble ofxgboost~\cite{chen2016xgboost} and TabPFN BO~\cite{tabpfn_bo_cookbook}, an extention of TabPFN that provides predictive distribution from TabPFN model as the two other modeling choices. For  these surrogate baselines, we incorporate both compute
window and fantasization.

The surrogate methods all recover the compute-optimal envelope faster than random search.
Random search reaches comparable coefficients but fails
to identify most of the compute-optimal points (poor envelope recovery) while still benefitting on
compute window. 
Note that both the dashed lines use only acquired points
to fit the scaling law, while the solid line uses
acquired and fantasized points on the grid.

\subsection{Generalizing to other scaling law datasets}\label{sec:app-datasets-base}

\begin{figure}[h!]
    \centering
    \includegraphics[width=\linewidth]{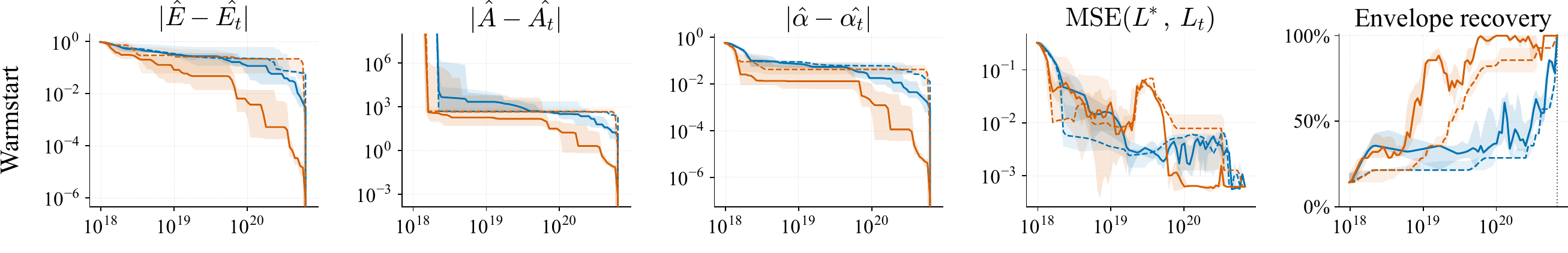}
    \includegraphics[width=\linewidth]{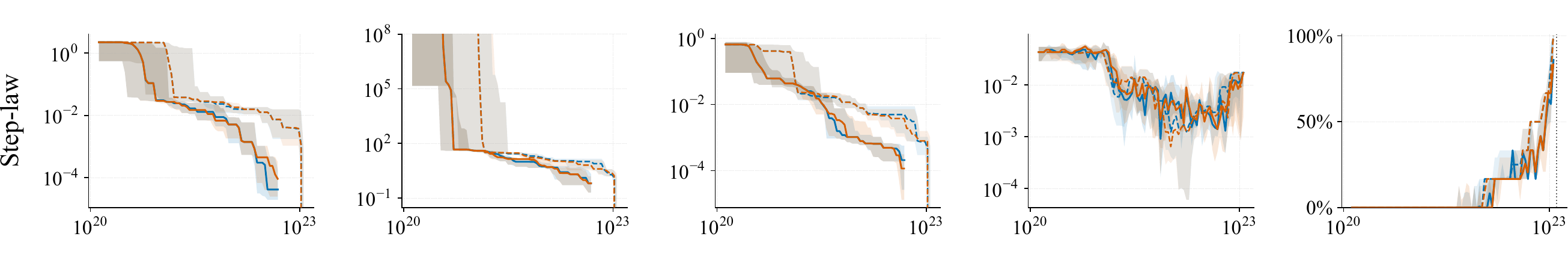}
        \includegraphics[width=\linewidth]{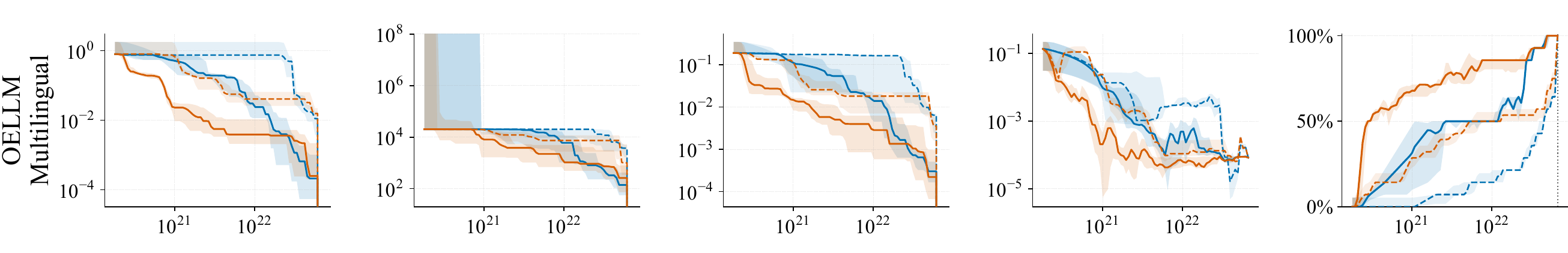}
    \includegraphics[width=\linewidth]{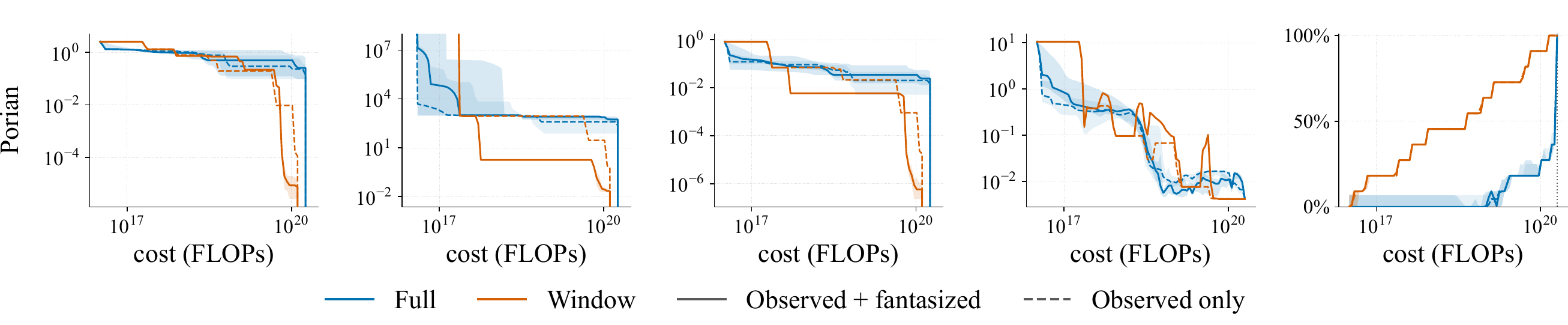}

    \caption{
    We provide $L(C)$ scaling law recovery plots
    on applying our approach to other scaling law studies (each row). Here we keep standard BO settings fixed: GP surrogate with Mat\'ern-5/2 kernel and EI as the acquisition function and show the results for plain (Full) and slabbed (Window) variants (dashed: only observed points used, solid: observed plus fantasized points used to construct $L(C)$).
    }
    \label{fig:app-generalized-datasets-lc}
\end{figure}

Next we apply the approach to other real-world scaling 
law benchmark datasets as defined in Table~\ref{tab:datasets_overview}. We show the results in Figure~\ref{fig:app-generalized-datasets-lc}. We note that while Warmstart, Porian and OELLM Multilingual studies
show improvement, Step-law is particularly poor in terms of recovering the envelope with our approach. We suspect the method fails to reliably recover the compute-optimal points for Step-law with less cost because the dataset contains many poor hyperparameter settings affecting the
surrogate.

Next we run some ablations on GP on the OELLM-English dataset on the $L(C)$ form. Unless otherwise specified, in
all plots we show the compute window plus fantasization
variants.

\subsection{GP kernel choices}\label{app:kernel choice}

\begin{figure}[h]
    \centering
    \includegraphics[width=\linewidth]{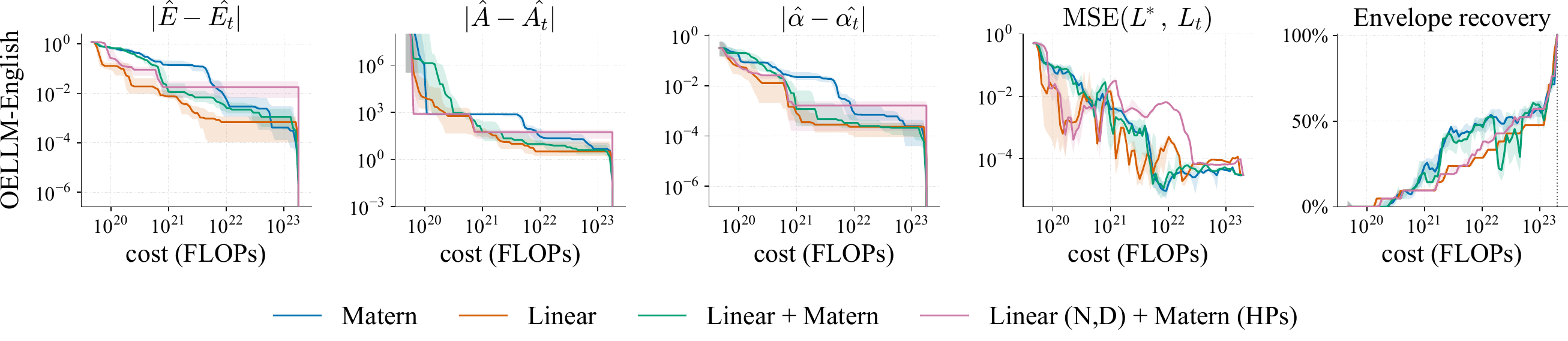}
    \caption{
    Different kernel choices for the GP surrogate with EI acquisition function. We experiment with $2$ standard kernels and $2$ additive kernels, with one applying different kernels to different variable types.
    }
    \label{fig:app-kernels-lc}
\end{figure}

Here we fix GP with EI acquisition, and vary the covariance kernel $k(x,x')$ of GP. We consider four kernels on the log-transformed input
$(\log N,\log D, \log\lambda)$ where $\lambda$ represents remaining hyperparameters. The default is a Mat\'ern-5/2 kernel $k_{5/2}$ on all
coordinates. Other kernels we consider are the linear
kernel $k_{\mathrm{lin}}$, sum of $k_{5/2}$ and $k_{\mathrm{lin}}$ on all coordinates, and a grouped additive form
$k_{\mathrm{lin}}(\log N, \log D)+k_{5/2}(\log \lambda)$ applies the linear term only to the
scale dimensions and the Mat\'ern term only to the hyperparameters.

We show the results in Figure~\ref{fig:app-kernels-lc}.
We see that while the Linear kernel often outperforms the other two on
parameter regrets, it is typically worse than the default
Mat\'ern or the Linear$+$Mat\'ern kernels. For this study, the grouped additive form also seem to not perform that
well across all tracked metrics.

\subsection{Choice of acquisition function}
\label{app:acquisition_function}
\begin{figure}[t]
    \centering
    \includegraphics[width=\linewidth]{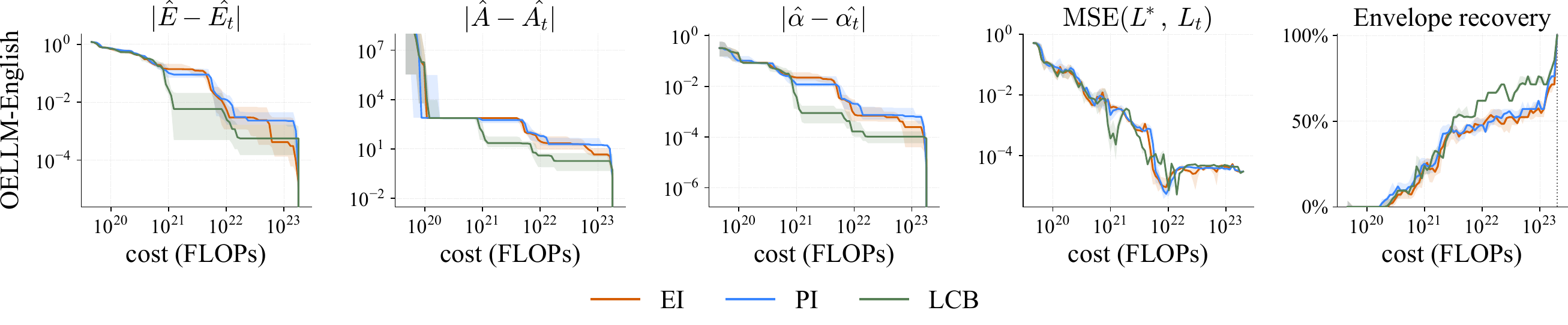}
    \caption{
    Three standard choices for the acquisition function used in BO. LCB performs slightly better than EI and PI on OELLM-English dataset.
    }
    \label{fig:app-af-lc}
\end{figure}

In Figure~\ref{fig:app-af-lc}, we ablate over the choice
of acquisition function used in the BO loop, and consider
3 standard choices: Expected Improvement (EI),
Probability of Improvement (PI) and Lower Confidence Bound (LCB)~\cite{kushner-jbe64a,mockus-ifip77a, garnett_bayesoptbook_2023}. We observe the GP+LCB combination performing
better than other two acquisition functions, and choose
to show the results for it in the main section of this study (Figure~\ref{fig:main_result}) and the mean incumbent coefficients recovered in Table~\ref{tab:lcb-incumbent-coeffs} at
different fractions of the total compute budget, complementing the Table \ref{tab:extrap-compute-lcb}.


\subsection{Parametric form: Chinchilla 3 Approach}\label{sec:app-chin}

We evaluate the Chinchilla 3 Approach $L(N, D)$ on the OELLM-English grid under the same surrogate-based acquisition protocol used for the $L(C)$ experiments described in Section \ref{sec:empirical}: 10 seeds (0--9), a Gaussian Process surrogate over $(N, D, \text{LR}, \text{BS})$, Expected Improvement selection, 1000 acquisition iterations, and 10 practice initialized points drawn from the full HP grid at the smallest $N$ over the four smallest $D$ values. Unlike the $L(C)$ experiment, which splits acquire-able pool and held-out set on compute, we split the grid on the model size axis and hold out the largest model size, which is never acquired. Each iteration scores the fitted law on the acquire-able pool (in-region recovery) and on the held-out large $N$ region (extrapolation to unseen scale).

\begin{figure}[h!]
    \centering
    \includegraphics[width=\linewidth]{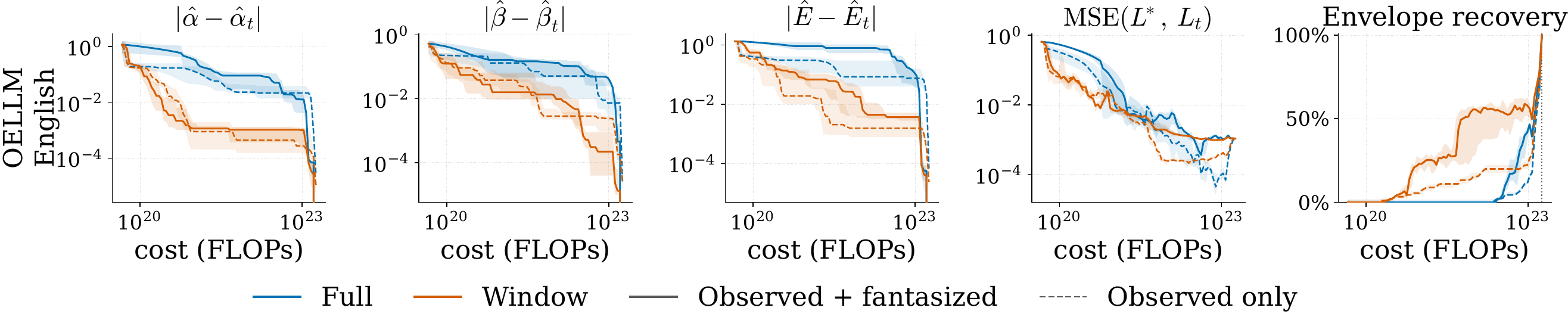}
    \vspace{-20pt}
    \caption{Recovery of the $L(N,D)$ parametric fit on OELLM-English under different BO variants. We compare acquisition over the full search space (\textit{Full}) and a growing compute window (\textit{Window}), with scaling laws fitted using either observed points only or observed and fantasized points. Panels report parameter recovery for $\alpha$, $\beta$, and $E$, held-out MSE, and recovery of the minimum-loss configuration at each $(N,D)$ pair. Shaded regions show $\pm1$ S.D. over 10 seeds.}
    \label{fig:chin_recovery}
\end{figure}

Figure \ref{fig:chin_recovery} shows that the conclusions obtained for $L(C)$ in Figure \ref{fig:main_result} extend to the two-dimensional $L(N,D)$ parametric form. Restricting acquisition to a growing compute window substantially speeds up the recovery of $\alpha$, $\beta$, and $E$, and achieves lower held-out prediction error at smaller cumulative compute. Fantasization is particularly beneficial for recovering the per-$(N,D)$ loss envelope early in the search. This result is notable because $L(N,D)$ imposes a stronger recovery requirement than $L(C)$: rather than identifying a one-dimensional compute-loss envelope, the acquisition procedure must obtain accurate minima across multiple $(N,D)$ pairs.

\subsection{Dataset experiments}\label{sec:chin-datasets}

\begin{figure}[h!]
    \centering
    
    \includegraphics[width=\linewidth]{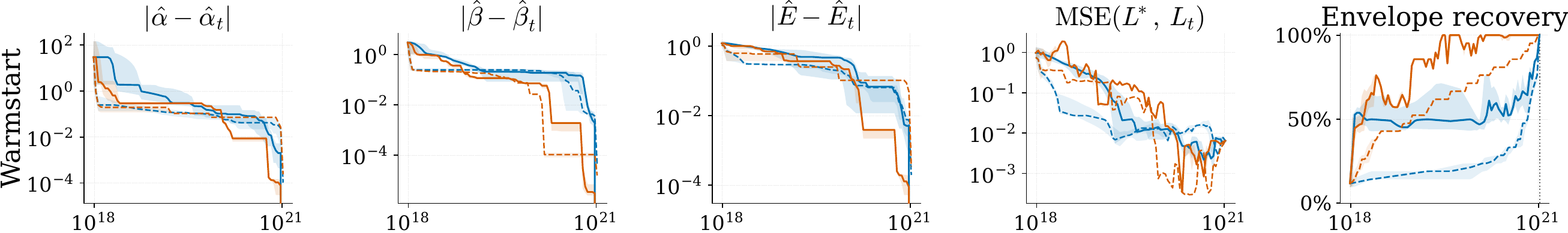}
    \includegraphics[width=\linewidth]{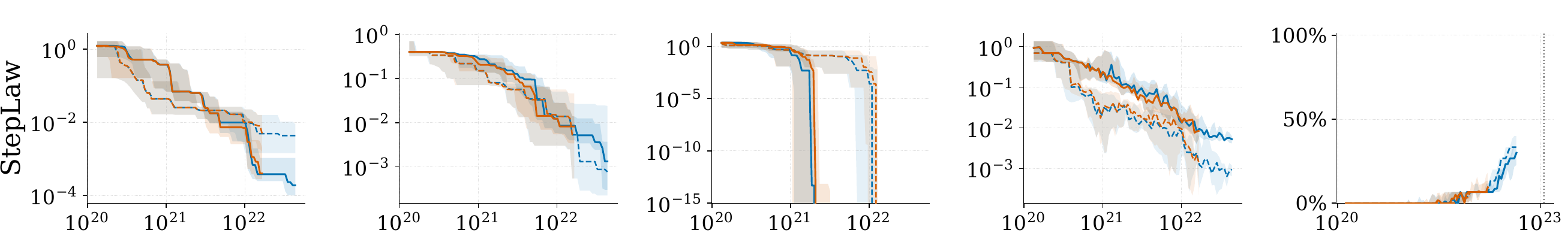}
    \includegraphics[width=\linewidth]{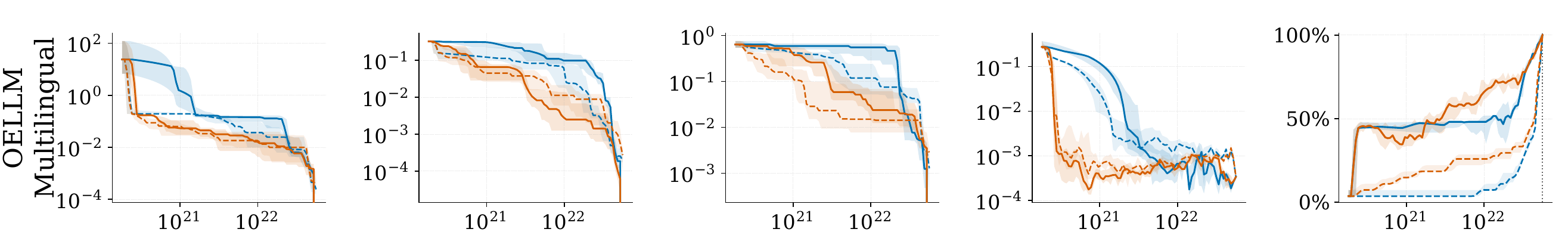}
    \includegraphics[width=\linewidth]{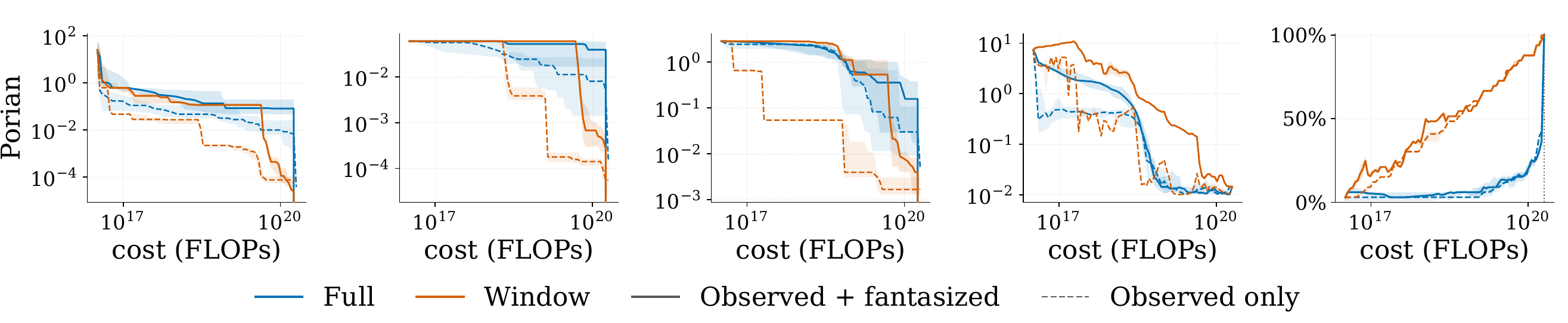}
    \caption{$L(N,D)$ scaling law recovery across benchmark datasets. Columns report recovery of $\alpha$, $\beta$, and $E$, held-out MSE, and per-$(N,D)$ envelope recovery. Lines compare \textit{Full} and \textit{Window} acquisition with observed-only and observed-plus-fantasized fits.}
    \label{fig:app-datasets-lnd}
\end{figure}

To assess whether the recovery behavior observed on OELLM-English generalizes across scaling study designs, we repeat the $L(N,D)$ experiment on the remaining benchmark datasets. We keep the BO configuration fixed and compare the same \textit{Full} and \textit{Window} acquisition variants, with and without fantasization. Figure \ref{fig:app-datasets-lnd} reports the resulting parameter recovery, held-out prediction error, and per-$(N,D)$ envelope recovery as a function of cumulative compute. The benefit of compute-window acquisition persists across datasets, although its magnitude depends on the underlying grid. The improvement is clearest for Warmstart, OELLM-Multilingual, and Porian, where the window generally recovers the fitted parameters and the per-$(N,D)$ loss envelope at lower cumulative compute. StepLaw remains substantially more challenging: the acquisition strategies show limited separation and recover only a small fraction of the loss envelope before the grid is nearly exhausted. Fantasization can further improve early envelope recovery, but its benefit is less consistent across datasets and individual fit metrics. Overall, these results indicate that compute slicing remains effective for the more demanding $L(N,D)$ recovery setting, while recovery efficiency is sensitive to the structure and density of the underlying experimental grid.

\end{document}